\pdfoutput=1

\documentclass[11pt]{article}

\usepackage{EMNLP2022}

\usepackage{times}
\usepackage{latexsym}
\usepackage{graphicx}
\usepackage{subfigure}
\usepackage{hyperref}
\usepackage{color}
\usepackage{booktabs} 

\usepackage[T1]{fontenc}

\usepackage[utf8]{inputenc}

\usepackage{microtype}

\usepackage{inconsolata}
\usepackage{soul}

\newcommand*{\changefontsql}{\fontfamily{qcr}\selectfont}
\newcommand{\sql}[1]{{\changefontsql{#1}}}
\newcommand*{\changefontprompt}{\fontfamily{cmtt}\selectfont}
\newcommand{\prompt}[1]{{\changefontprompt{#1}}}
\newcommand{\dataset}[1]{\texttt{#1}}
\newcommand{\red}[1]{\textcolor{red}{#1}}

\newcommand{\blue}[1]{\textcolor{blue}{#1}}

\title{Towards Generalizable and Robust Text-to-SQL Parsing\thanks{\hspace{1mm} Work done when Chang Gao was an intern at Alibaba. The work described in this paper is substantially supported by a grant from the Research Grant Council of the Hong Kong Special Administrative Region, China (Project Code: 14204418).}}

\author{Chang Gao\textsuperscript{\rm 1}, Bowen Li\textsuperscript{\rm 2}, Wenxuan Zhang\textsuperscript{\rm 2 }, Wai Lam\textsuperscript{\rm 1\footnotemark[2]}, Binhua Li\textsuperscript{\rm 2}, \\ {\bf Fei Huang\textsuperscript{\rm 2}, Luo Si\textsuperscript{\rm 2} and Yongbin Li\textsuperscript{\rm 2\footnotemark[2]}} \\
\textsuperscript{\rm 1}The Chinese University of Hong Kong \\ \textsuperscript{\rm 2}DAMO Academy, Alibaba Group \\
\texttt{\{gaochang,wlam\}@se.cuhk.edu.hk, libowen.ne@gmail.com}  \\
\texttt{\{saike.zwx,binhua.lbh,shuide.lyb\}@alibaba-inc.com} } 

\begin{document}
\maketitle
\renewcommand{\thefootnote}{\fnsymbol{footnote}}
\footnotetext[2]{Corresponding authors.}
\renewcommand{\thefootnote}{\arabic{footnote}}
\begin{abstract}

Text-to-SQL parsing tackles the problem of mapping natural language questions to executable SQL queries. In practice, text-to-SQL parsers often encounter various challenging scenarios, requiring them to be generalizable and robust. While most existing work addresses a particular generalization or robustness challenge, we aim to study it in a more comprehensive manner. In specific, we believe that text-to-SQL parsers should be (1) \textbf{generalizable} at three levels of generalization, namely \textit{i.i.d.}, \textit{zero-shot}, and \textit{compositional}, and (2) \textbf{robust} against input perturbations. To enhance these capabilities of the parser, we propose a novel TKK framework consisting of Task decomposition, Knowledge acquisition, and Knowledge composition to learn text-to-SQL parsing in stages. By dividing the learning process into multiple stages, our framework improves the parser's ability to acquire general SQL knowledge instead of capturing spurious patterns, making it more generalizable and robust. Experimental results under various generalization and robustness settings show that our framework is effective in all scenarios and achieves state-of-the-art performance on the Spider, SParC, and CoSQL datasets. Code can be found at \url{https://github.com/AlibabaResearch/DAMO-ConvAI/tree/main/tkk}.

\end{abstract}

\section{Introduction}
\label{intro}

\begin{figure*}[tb]
  \centering
  \includegraphics[width=0.9\linewidth]{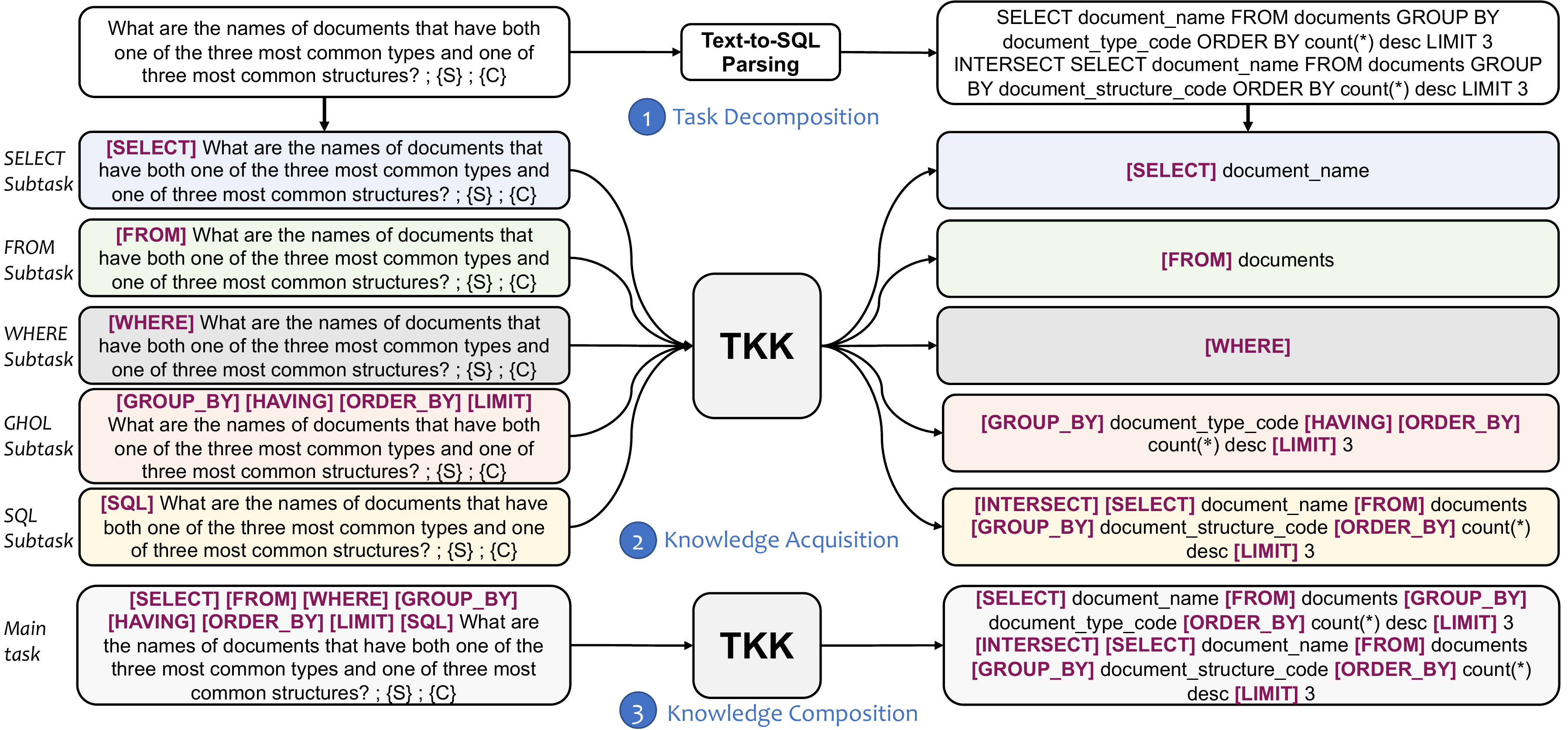}  
  \caption{Overview of our TKK framework. \{S\} and \{C\} denote the database schema and context, respectively.}
  \label{fig:model}
\end{figure*}

Text-to-SQL parsing aims to translate natural language questions to SQL queries that can be executed on databases to produce answers \cite{lin-etal-2020-bridging}, which bridges the gap between expert programmers and ordinary users who are not proficient in writing SQL queries. Thus, it has drawn great attention in recent years \cite{zhong2017seq2sql, suhr-etal-2020-exploring, picard, s2sql, qin2022survey, qin-etal-2022-sun}. 

Early work in this field  \cite{zelle1996learning, yaghmazadeh2017sqlizer, iyer-etal-2017-learning} mainly focuses on \textit{i.i.d. generalization}. They only use a single database, and the exact same target SQL query may appear in both the training and test sets.  However, it is difficult to collect sufficient training data to cover all the questions users may ask \cite{Beyond} and the predictions of test examples might be obtained by semantic matching instead of semantic parsing \cite{spider}, limiting the generalization ability of parsers. Subsequent work further focuses on generalizable text-to-SQL parsing in terms of two aspects: \textit{zero-shot generalization} and \textit{compositional generalization}. Zero-shot generalization requires the parser to generalize to unseen database schemas.
Thanks to large-scale datasets such as Spider \cite{spider}, SParC \cite{sparc}, and CoSQL \cite{cosql}, zero-shot generalization has been the most popular setting for text-to-SQL parsing in recent years.
Various methods involving designing graph-based encoders \cite{ratsql, lgesql} and syntax tree decoders \cite{syntaxsqlnet, smbop} have been developed to tackle this challenge.
Compositional generalization is the desired ability to generalize to test examples consisting of novel combinations of components observed during training. \citet{finegan-dollak-etal-2018-improving} explore compositional generalization in text-to-SQL parsing focusing on template-based query splits. \citet{shaw-etal-2021-compositional} provide new splits of Spider considering length, query template, and query compound divergence to create challenging evaluations of compositional generalization.

Another challenge of conducting text-to-SQL parsing in practice is \textit{robustness}. Existing text-to-SQL models have been found vulnerable to input perturbations  \cite{deng-etal-2021-structure, gan-etal-2021-towards, pi-etal-2022-towards}. For example, \citet{gan-etal-2021-towards} replace schema-related words in natural language questions with manually selected synonyms and observe a dramatic performance drop. They propose two approaches, namely multi-annotation selection and adversarial training, to improve model robustness against synonym substitution.

Although specialized model architectures and training approaches have been proposed to address a particular generalization or robustness challenge, we believe that practical text-to-SQL parsers should be built with strong generalizability in terms of \textbf{all three levels of generalization}, namely \textit{i.i.d.}, \textit{zero-shot}, and \textit{compositional}, and \textbf{robustness} against input perturbations.
To obtain such capabilities, it can be noticed that humans often learn to write each clause, such as \sql{SELECT} or \sql{WHERE}, for a basic operation, before composing them to fulfill a more challenging goal, i.e., writing the entire SQL query. In contrast, most existing methods adopt a one-stage learning paradigm, i.e., learning to write each SQL clause and the dependency between different clauses simultaneously. This may lead the model to capture spurious patterns between the question, database schema, and SQL query instead of learning general SQL knowledge.

To this end, we propose a novel framework consisting of three learning stages including \textit{\textbf{T}ask decomposition, \textbf{K}nowledge acquisition, and \textbf{K}nowledge composition} (TKK) for text-to-SQL parsing, which mimics the human learning procedure to learn to handle the task in stages. Specifically, in the task decomposition stage, TKK decomposes the original task into several subtasks. Each subtask corresponds to mapping the natural language question to one or more clauses of the SQL query, as shown in the top portion of Figure \ref{fig:model}. 
Afterwards, TKK features a prompt-based learning strategy to separately acquire the knowledge of subtasks and employ the learned knowledge to tackle the main task, i.e., generating the entire SQL query. 
In the knowledge acquisition stage, TKK trains the model with all the subtasks in a multi-task learning manner; 
in the knowledge composition stage, TKK fine-tunes the model with the main task to combine the acquired knowledge of subtasks and learn the dependency between them. 

The advantages of our three-stage framework over previous one-stage learning methods are three-fold: (1) it reduces the difficulty of model learning by dividing the learning process into multiple easier-to-learn stages; (2) it explicitly forces the model to learn the alignment between the question, database schema, and each SQL clause as it needs to identify the intent expressed in the question based on the schema to generate a specific clause; (3) by explicitly constructing the training data for each subtask, it is easier for the model to learn the knowledge required to translate the question into each SQL clause. These advantages help the model to learn general SQL knowledge rather than some dataset-specific patterns, making it more generalizable and robust.

To verify the effectiveness of our framework, we conduct comprehensive evaluations on representative benchmarks covering all three levels of generalization and robustness scenarios with pre-trained sequence-to-sequence models. Experimental results and analysis show that: (1) we achieve state-of-the-art performance on the Spider, SParC, and CoSQL datasets; (2) our method outperforms vanilla sequence-to-sequence models in all  scenarios; (3) our framework significantly improves the model's ability to generate complex SQL queries; (4) our framework is also effective in the low-resource setting.

\section{Background}
\textbf{Notations} 
We use the lowercase letter $q$ to denote a natural language question and denote its corresponding database schema, context, and SQL query as $s_q$, $c_q$, and $l_q$, respectively. We represent the set of training examples $(q, s_q, c_q, l_q)$ as $\mathcal{D}_{train}$ and test set as $\mathcal{D}_{test}$. A perturbed test set $\mathcal{D^\prime}_{test}$ could be constructed by perturbations to questions such as synonym substitution  to form $(q^\prime, s_{q}, c_{q}, l_{q})$. 
We denote $\mathcal{S}_{train}$ as the set of database schemas of $\mathcal{D}_{train}$, $\mathcal{L}_{train}$ as the set of SQL queries of $\mathcal{D}_{train}$, and $\mathcal{Q}_{test}$ as the set of questions of  $\mathcal{D}_{test}$.
\\
\\
\textbf{Problem Definition} Given  $(q, s_q, c_q)$, where the database schema $s_q$ consists of tables and columns, and context $c_q$ is the interaction history consisting of previous questions and system clarification in the multi-turn setting or empty in the single-turn setting, the goal is to generate the correct SQL query $l_q$. 
\\
\\
\textbf{Generalization and Robustness}
Following \citet{Beyond} and \citet{robustness}, we formalize three levels of generalization and robustness as follows:

\textit{Zero-shot generalization}: $\forall q\in \mathcal{Q}_{test},\ s_q\not\in  \mathcal{S}_{train}$.

\textit{Compositional generalization}: $\forall q\in \mathcal{Q}_{test}, \ s_q\in \mathcal{S}_{train}, \ l_q\not\in \mathcal{L}_{train}$.

\textit{I.I.D. generalization}: $\forall q\in \mathcal{Q}_{test}, \ s_q\in \mathcal{S}_{train}$.  $\mathcal{D}_{train}$ and $\mathcal{D}_{test}$ follow the same distribution.

\textit{Robustness}: training with $\mathcal{D}_{train}$ but adopting $\mathcal{D^\prime}_{test}$ instead of $\mathcal{D}_{test}$ for evaluation.

\section{Our TKK Framework}

TKK consists of three learning stages: task decomposition, knowledge acquisition, and knowledge composition. In this section, we first introduce each stage in detail. Then we describe the training and inference of TKK.

\subsection{Three Stages of TKK}

\textbf{Task Decomposition}
As shown in Figure \ref{fig:model}, we decompose the text-to-SQL parsing task into five subtasks, namely \sql{SELECT}, \sql{FROM}, \sql{WHERE}, \sql{GHOL}, and \sql{SQL}. Basically, a subtask aims to translate the natural language question to one or more clauses of the SQL query. For example, the \sql{GHOL} subtask aims to generate the the \sql{GROUP\_BY}, \sql{HAVING}, \sql{ORDER\_BY}, and \sql{LIMIT} clauses given the question and its corresponding database schema and context. For queries involving set operators such as \sql{INTERSECT}, \sql{UNION}, and \sql{EXCEPT} to combine two SQL queries, we treat the first query as usual and the second query as the \sql{SQL} clause of the first query. The \sql{SQL} subtask targets at mapping the question to the \sql{SQL} clause. 

There are two considerations behind constructing a subtask: (1) the number of classification examples; (2) the dependency between different clauses. First, according to the SQL syntax, every SQL has the \sql{SELECT} and \sql{FROM} clauses. However, clauses such as \sql{GROUP\_BY} and \sql{ORDER\_BY} appear only in relatively complicated SQL queries. It implies that the number of these clauses is much smaller than that of the \sql{SELECT} or \sql{FROM} clause. Trivially considering generating each clause as a subtask is problematic. If a specific clause does not exist, the generation task degenerates to a classification task because the model only needs to judge its existence. We denote these examples as \textit{classification examples}. Too many classification examples are harmful to model learning. Second, the \sql{GROUP\_BY} and \sql{HAVING} clauses are usually bundled together, which is also the case of the \sql{ORDER\_BY} and \sql{LIMIT} clauses. The \sql{ORDER\_BY} clause is often dependent on the \sql{GROUP\_BY} clause if they appear in a SQL query simultaneously.
Based on the above observations, combining these clauses to construct a single subtask is more appropriate. 
We do not further decompose the \sql{SQL} clause because there will be more subtasks, and most training examples of these subtasks are classification examples. 
\\
\\
\textbf{Knowledge Acquisition}
In this stage, we train the sequence-to-sequence model with all subtasks using multi-task learning.  We assign each SQL keyword a special token, which is also used to denote its corresponding clause. Then we construct a task prompt for each subtask based on the clauses it contains. For example, the special token corresponding to \sql{GROUP\_BY} is ``[\prompt{GROUP\_BY}]'' and the prompt for the \sql{GHOL} subtask is ``[\prompt{GROUP\_BY}] [\prompt{HAVING}] [\prompt{ORDER\_BY}] [\prompt{LIMIT}]''. The input for each subtask simply adds a task prompt to the input for the original task.

For constructing the target, we replace the keywords in each clause with their corresponding special tokens. If a clause is empty, we use its corresponding special token to build the target. For instance, the example in Figure \ref{fig:model} does not contain the \sql{WHERE} clause. Thus the target of the \sql{WHERE} subtask is ``[\prompt{WHERE}]''. Those examples whose targets only contain special tokens are \textit{classification examples}, as we mentioned earlier. For those examples whose targets contain at least one non-empty clause, we regard them as \textit{parsing examples}. Classification examples are helpful since the model needs to learn which clauses to generate given a particular question. However, too many classification examples make it difficult for the model to learn the knowledge of subtasks. Even though we pack the \sql{GROUP\_BY}, \sql{HAVING}, \sql{ORDER\_BY}, and \sql{LIMIT} clauses into one subtask, the number of classification examples is still much bigger than that of parsing examples. The \sql{SQL} subtask  also has the problem. To tackle this problem, we downsample classification examples for each subtask to guarantee that the proportion of parsing examples is at least a ratio $r$.  
\\
\\
\textbf{Knowledge Composition}
Training the model with multiple subtasks cannot capture the interdependency between them. In this stage, we fine-tune the model with the main task, i.e., generating the entire SQL query, to capture such information. As shown in Figure \ref{fig:model}, we combine the prompts of subtasks to construct the prompt of the main task to guide the model to composite the knowledge of subtasks. 

\subsection{Training and Inference}

We formulate text-to-SQL parsing as a sequence-to-sequence generation problem. The input is the serialization of the question, database schema, and context, and the output is the SQL query. In the knowledge acquisition and composition stages, we adjust the input and output according to what we discussed in the last section. We adopt the pre-trained sequence-to-sequence model T5 \cite{T5} as the backbone of TKK.
\\
\\
\textbf{Training}
The model is trained with a maximum likelihood objective. Given the training example $(q,s_q,c_q,tp,y)$, the loss function $L_\theta$ is defined as
\begin{equation}
L_{\theta}=-\sum_{i=1}^{n} \log P_{\theta}\left(y_{i} \mid y_{<i}, q, s_q, c_q, tp\right)
\end{equation}
where $\theta$ is the model parameters, $tp$ is the task prompt, $y$ is the target sequence, and $n$ is the length of $y$. In the knowledge acquisition stage, we mix the data of all subtasks for training. In the knowledge composition stage, we initialize the model with the weights of the model trained in the knowledge acquisition stage and use the data of the main task for training.
\\
\\
\textbf{Inference}
After training, for each triple of the question, database schema, and context $(q,s_q,c_q)$, we generate the target sequence of the main task for obtaining the SQL query. We replace the special tokens in the target sequence with their corresponding SQL keywords.

\section{Experiments}

\subsection{Experimental Setup}

\begin{table}
\centering
\setlength\tabcolsep{3.5pt}
\begin{tabular}{lcc}
\hline
\textbf{Models} & \textbf{EM} & \textbf{EX} \\
\hline
 \small{Global-GNN \cite{global-gnn}} & 52.7 & - \\
 \small{IRNet + BERT \cite{irnet}}  & 63.9 & - \\
 \small{RATSQL + BERT \cite{ratsql}} & 69.7 & - \\
 \small{RYANSQL + BERT \cite{ryansql}} & 70.6 & - \\
 \small{RATSQL + GraPPa \cite{yu2021grappa}} & 73.4 & - \\
 \small{LGESQL + ELECTRA \cite{lgesql}} & 75.1 & - \\
\small{T5-Base\dag \  \cite{T5}} & 58.1 & 60.1 \\
\small{T5-Large\dag \ \cite{T5}} & 66.6 & 68.3 \\
\small{T5-3B\dag \ \cite{T5}}  & 71.8 & 74.4 \\
\small{T5-Base + PICARD \cite{picard} }  & 65.8 & 68.4 \\
\small{T5-Large + PICARD \cite{picard}}  & 69.1 & 72.9 \\
\small{T5-3B + PICARD \cite{picard} } & 75.5 & 79.3 \\
\hline
\small{TKK-Base} & 61.5 & 64.2 \\
\small{TKK-Large} & 70.6 & 73.2 \\
\small{TKK-3B} & 74.2 & 78.4 \\
\small{TKK-Base + PICARD} & 70.4  &  76.0 \\
\small{TKK-Large + PICARD} & 74.1 & 78.2 \\
\small{TKK-3B + PICARD} & \textbf{75.6} & \textbf{80.3} \\
\hline
\end{tabular}
\caption{Zero-shot generalization results on \dataset{Spider}. [\dag]: Results are taken from \cite{UnifiedSKG}.}
\label{tab:spider}
\end{table}

\begin{table*}
\centering
\begin{tabular}{lcccc}
\hline
&  \multicolumn{2}{c}{\dataset{SParC}} & \multicolumn{2}{c}{\dataset{CoSQL}} \\
\textbf{Models} & \textbf{QM} & \textbf{IM} & \textbf{QM} & \textbf{IM} \\
\hline
EditSQL + BERT \cite{editsql} & 47.2 & 29.5 & 39.9 & 12.3 \\
IGSQL + BERT \cite{igsql} & 50.7 & 32.5 & 44.1 & 15.8 \\
R$^2$SQL + BERT \cite{r2sql} & 54.1 & 35.2 & 45.7 & 19.5 \\
RAT-SQL + SCoRe \cite{yu2021score} & 62.2 & 42.5 & 52.1 & 22.0 \\
HIE-SQL + GraPPa \cite{hiesql} & 64.7 & 45.0  & 56.4 & \textbf{28.7} \\
T5-Base\dag \ \cite{T5}  & 50.6 & 31.3 & 42.3 & 12.6\\
T5-Large\dag \ \cite{T5} & 56.7 & 37.4 & 48.3 & 16.7 \\
T5-3B\dag \ \cite{T5}  & 61.5 & 41.9 & 54.1 & 22.8 \\
T5-3B + PICARD \cite{picard} & - & - & 56.9 & 24.2 \\
\hline
TKK-Base & 52.6 & 32.7 & 46.9 & 17.8 \\
TKK-Large & 60.2 & 41.0 & 50.5 & 21.5 \\
TKK-3B & 65.5 & 46.7 & 54.9 & 24.9 \\
TKK-3B + PICARD & \textbf{66.6} & \textbf{48.3} & \textbf{58.3} & 27.3 \\
\hline
\end{tabular}
\caption{Zero-shot generalization results on \dataset{SParC} and \dataset{CoSQL}. [\dag]: Results are taken from \cite{UnifiedSKG}.}
\label{tab:sparc+cosql}
\vspace{-0.2cm}
\end{table*}

\begin{table*}
\centering
\begin{tabular}{lcccccc}
\hline
& \multicolumn{2}{l}{\dataset{Spider-Template}} & \multicolumn{2}{l}{\dataset{Spider-Length}} &  \multicolumn{2}{l}{\dataset{Spider-TMCD}} \\
\textbf{Models} & \textbf{EM} & \textbf{EX} & \textbf{EM} & \textbf{EX} & \textbf{EM} & \textbf{EX}\\
\hline
T5-Base\dag \ \cite{T5}  & 59.3 & - & 49.0 & - & 60.9 & - \\
T5-3B\dag \ \cite{T5} & 64.8 & - & 56.7 & - & 69.6 & - \\
NQG-T5-3B \cite{shaw-etal-2021-compositional} & 64.7 & - & 56.7 & - & 69.5 & - \\
\hline
TKK-Base & 62.9 & 69.8 & 52.0 & 55.3 & 63.3 & 71.3 \\
TKK-3B & \textbf{70.3} & \textbf{77.2} & \textbf{58.6} & \textbf{63.3} & \textbf{71.8} & \textbf{79.1} \\
\hline
\end{tabular}
\caption{Results on three compositional splits of \dataset{Spider}. [\dag]: Results are taken from \cite{shaw-etal-2021-compositional}.}
\label{tab:spider-comp}
\end{table*}

\textbf{Datasets}
For \textit{zero-shot generalization}, we use the original \dataset{Spider} \cite{spider}, \dataset{SParC} \cite{sparc}, and \dataset{CoSQL} \cite{cosql} datasets\footnote{Since the test sets of these datasets are not public, we report results on the development sets.}. \dataset{Spider} is a single-turn dataset, while \dataset{SParC} and \dataset{CoSQL} are multi-turn datasets. For \textit{compositional generalization}, we use three compositional splits derived from \dataset{Spider}, namely template split (\dataset{Spider-Template}), length split (\dataset{Spider-Length}), and Target Maximum Compound Divergence (TMCD) split  (\dataset{Spider-TMCD}), from \citet{shaw-etal-2021-compositional}.
For \textit{i.i.d. generalization}, we construct \dataset{Spider-IID}, \dataset{SParC-IID}, and \dataset{CoSQL-IID} based on \dataset{Spider}, \dataset{SParC}, and \dataset{CoSQL}, respectively. For example, to obtain \dataset{Spider-IID}, we mix the training and development sets of \dataset{Spider} to get the full set and then randomly sample from it to construct new training and development sets while retaining the ratio of the number of original training and development examples. 

For \textit{robustness}, we use \texttt{Spider-Syn} \cite{gan-etal-2021-towards} and \texttt{Spider-Realistic} \cite{deng-etal-2021-structure} for evaluation. \texttt{Spider-Syn} is constructed via modifying questions in \dataset{Spider} using synonym substitution. 
\texttt{Spider-Realistic} selects a complex subset from the development set of \dataset{Spider} and modifies the questions in this subset to remove or paraphrase explicit mentions of column names while keeping the SQL queries unchanged. 
\\
\\
\textbf{Evaluation Metrics}
For \dataset{Spider} and datasets derived from it, we use Exact Match (EM) and Execution Accuracy (EX) following \citet{spider}. For \dataset{SParC}, \dataset{CoSQL}, and datasets derived from them, we use Question Match (QM) and Interaction Match (IM) following \citet{sparc}.
\\
\\
\textbf{Implementation Details}
TKK has three model sizes: TKK-Base, TKK-Large, and TKK-3B, which are initialized with pre-trained T5-Base, T5-Large, and T5-3B models \cite{T5}, respectively. We use the same Question-Schema-Context serialization as in \cite{UnifiedSKG}. We set the maximum input length to 512, the maximum target length to 128, and the batch size to 32. We use the Adafactor \cite{Adafactor} optimizer for all experiments. We set the learning rate to 1e-4 for TKK-Base and TKK-Large and 5e-5 for TKK-3B and use linear learning rate decay. In addition, we choose the data balance ratio $r$ from \{0.5, 0.7, 0.9\}.
All experiments are done on NVIDIA Tesla A100 and V100.

\begin{table*}
\centering
\begin{tabular}{lcccccc}
\hline
& \multicolumn{2}{l}{\dataset{Spider-IID}} & \multicolumn{2}{l}{\dataset{SParC-IID}} &  \multicolumn{2}{l}{\dataset{CoSQL-IID}} \\
\textbf{Models} & \textbf{EM} & \textbf{EX} & \textbf{QM} & \textbf{IM} & \textbf{QM} & \textbf{IM}\\
\hline
T5-Base  \cite{T5}  & 84.1 & 86.2 & 68.3 & 44.6 & 47.9 & 17.8 \\
T5-Large \cite{T5} & 86.9 & 88.5 & 70.0 & 49.1 & 52.9 & 23.6 \\
\hline
TKK-Base & 86.6 & 88.1 & 70.3 & 46.9 & 51.5 & 22.2 \\
TKK-Large & \textbf{88.3} & \textbf{89.8} & \textbf{72.3} & \textbf{52.6} & \textbf{56.9} & \textbf{27.3} \\
\hline
\end{tabular}
\caption{Results on three datasets for i.i.d. generalization: \dataset{Spider-IID}, \dataset{SParC-IID}, and \dataset{CoSQL-IID}.}
\label{tab:iid}
\end{table*}

\subsection{Generalization Results}

Tables \ref{tab:spider} and \ref{tab:sparc+cosql} report the zero-shot generalization results on \dataset{Spider}, \dataset{SParC}, and \dataset{CoSQL}, respectively. Equipped with PICARD \cite{picard}, which constrains the decoders to generate valid SQL queries by rejecting inadmissible tokens, TKK-3B achieves state-of-the-art results on these three datasets, demonstrating the strong \textit{zero-shot generalization} ability of our framework. Noticeably, TKK outperforms T5 on all datasets and all model sizes. 
Zero-shot generalization is challenging as it requires the model to accurately understand a question conditioned on an unseen database schema to generate the correct SQL query. As a result, the model has to acquire general SQL knowledge rather than trivially memorize seen SQL patterns. Our framework forces the model to align the question, database schema, and each SQL clause and helps the model to learn SQL knowledge, thus leading to better generalization performance. In addition, as shown in Table \ref{tab:spider}, \mbox{TKK-Base} with PICARD achieves comparable performance to strong specialized models such as RYANSQL, indicating the great potential of pre-trained sequence-to-sequence models for text-to-SQL parsing. Note that previous state-of-the-art models such as LGESQL and HIE-SQL heavily rely on manual design and may overfit to specific datasets. On the contrary, our framework enjoys strong generality as well as effectiveness. 

Table \ref{tab:spider-comp} presents the results on the three compositional splits of \dataset{Spider}. 
TKK outperforms T5 on all the three splits, demonstrating its powerful \textit{compositional generalization} ability. By comparison, NQG-T5, which combines a grammar-based approach NQG with T5, shows no gain over T5 on these datasets.
\dataset{Spider-Template} and \dataset{Spider-TMCD} require the model to generalize to novel templates and atom combinations, respectively, while \dataset{Spider-Length} requires the model to generalize to longer outputs. By explicitly decomposing the original task into multiple subtasks and combining the knowledge of them, our framework enables the model to better learn SQL knowledge and makes it less sensitive to these changes.

Table \ref{tab:iid} shows the results of TKK and the strong baseline T5 model on \dataset{Spider-IID}, \dataset{SParC-IID}, and \dataset{CoSQL-IID}. TKK obtains better results than T5 on all three datasets, demonstrating our framework's strong \textit{i.i.d. generalization} ability. It can be seen that \textit{i.i.d. generalization} is not as challenging as the other two generalization scenarios. However, the results on \dataset{SParC-IID} and \dataset{CoSQL-IID} are still not satisfactory. Enhancing the model's ability to acquire general knowledge is also helpful and necessary in this setting.

\subsection{Robustness Results}

Table \ref{tab:robust} reports the results of various models trained on \dataset{Spider} and evaluated on \dataset{Spider}, \dataset{Spider-Syn}, and \dataset{Spider-Realistic}, which measures the model's robustness against perturbations to natural language questions.
We have the following observations: (1) T5 is more robust than specialized models. For example, when evaluated on \dataset{Spider-Syn}, RATSQL + BERT degrades by 21.5 absolute points on EM, while T5-3B sees a performance drop of 12.2 absolute points. T5-Large, which performs worse than  RATSQL + BERT on \dataset{Spider}, can obtain better performance than it on \dataset{Spider-Syn}. This indicates that models specially designed for \dataset{Spider} are prone to overfitting on it. Thus evaluating their robustness is important. (2) Our TKK framework can improve the robustness of T5 for text-to-SQL parsing. TKK outperforms T5 on both \dataset{Spider-Syn} and \dataset{Spider-Realistic} for all model sizes. (3) STRUG improves robustness via structure-grounded pre-training with a large-scale of text-table paired data, while TKK provides a better way for learning text-to-SQL parsing to achieve this and does not need any additional data. 
(4) For pre-trained sequence-to-sequence models, the larger the model is, the more robust it is. When the model becomes larger, the gap between the performance of TKK on \dataset{Spider-Syn} and \dataset{Spider-Realistic} and the performance on \dataset{Spider} narrows. The same trend can be seen with T5.

\begin{table*}
\centering
\begin{tabular}{lcccccc}
\hline
& \multicolumn{2}{c}{\dataset{Spider}} & \multicolumn{2}{l}{\dataset{Spider-Syn}} &  \multicolumn{2}{l}{\dataset{Spider-Realistic}} \\
\textbf{Models} & \textbf{EM} & \textbf{EX} & \textbf{EM} & \textbf{EX} & \textbf{EM} & \textbf{EX}\\
\hline
IRNet \cite{irnet} & 53.2 & - & 28.4 & - & - & - \\
RAT-SQL + BERT \cite{ratsql} & 69.7 & - & 48.2 & - & 58.1 & 62.1 \\
RAT-SQL + STRUG \cite{deng-etal-2021-structure} & 72.6 & 74.9 & - & - & 62.2 & 65.3 \\
T5-Base\dag \ \cite{T5}  & 56.8 & 59.9 & 40.8 & 43.8 & 46.9 & 47.6 \\
T5-Large\dag \ \cite{T5} & 66.8 & 70.9 & 53.1 & 57.4 & 57.7 & 60.0 \\
T5-3B\dag \ \cite{T5} & 71.6 & 74.5 & 59.4 & 65.3 & 63.2 & 65.0 \\
\hline
TKK-Base & 61.5 & 64.2 & 44.2 & 47.7 & 53.7 & 53.7 \\
TKK-Large & 70.6 & 73.2 & 55.1 & 60.5 & 64.4 & 64.4 \\
TKK-3B & \textbf{74.2} & \textbf{78.4} & \textbf{63.0} & \textbf{68.2} & \textbf{68.5} & \textbf{71.1} \\
\hline
\end{tabular}
\caption{Results of models trained on \dataset{Spider} and evaluated on \dataset{Spider}, \dataset{Spider-Syn} and \dataset{Spider-Realistic}. [\dag]: We train T5 models on \dataset{Spider} and report evaluated results on the three datasets, which are different from Table \ref{tab:spider}.}
\label{tab:robust}
\end{table*}

\section{More Analysis}

\begin{figure}[tb]
\centering
\subfigure[EM results on \dataset{Spider}]{
\centering
\begin{minipage}[t]{0.5\linewidth}
\centering
\includegraphics[width=1.4in]{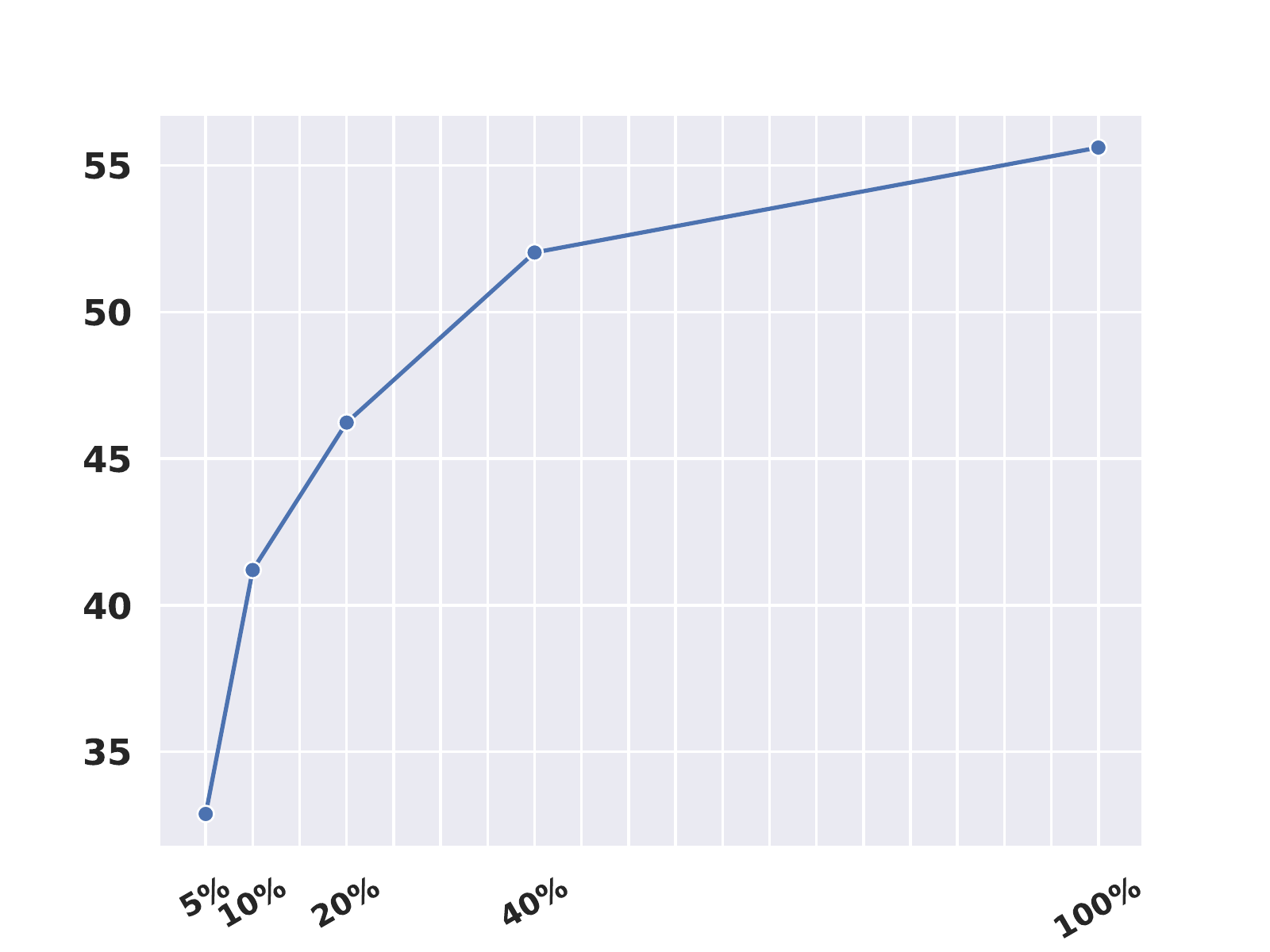}
\end{minipage}%
}%
\subfigure[QM results on \dataset{CoSQL}]{
\centering
\begin{minipage}[t]{0.5\linewidth}
\centering
\includegraphics[width=1.4in]{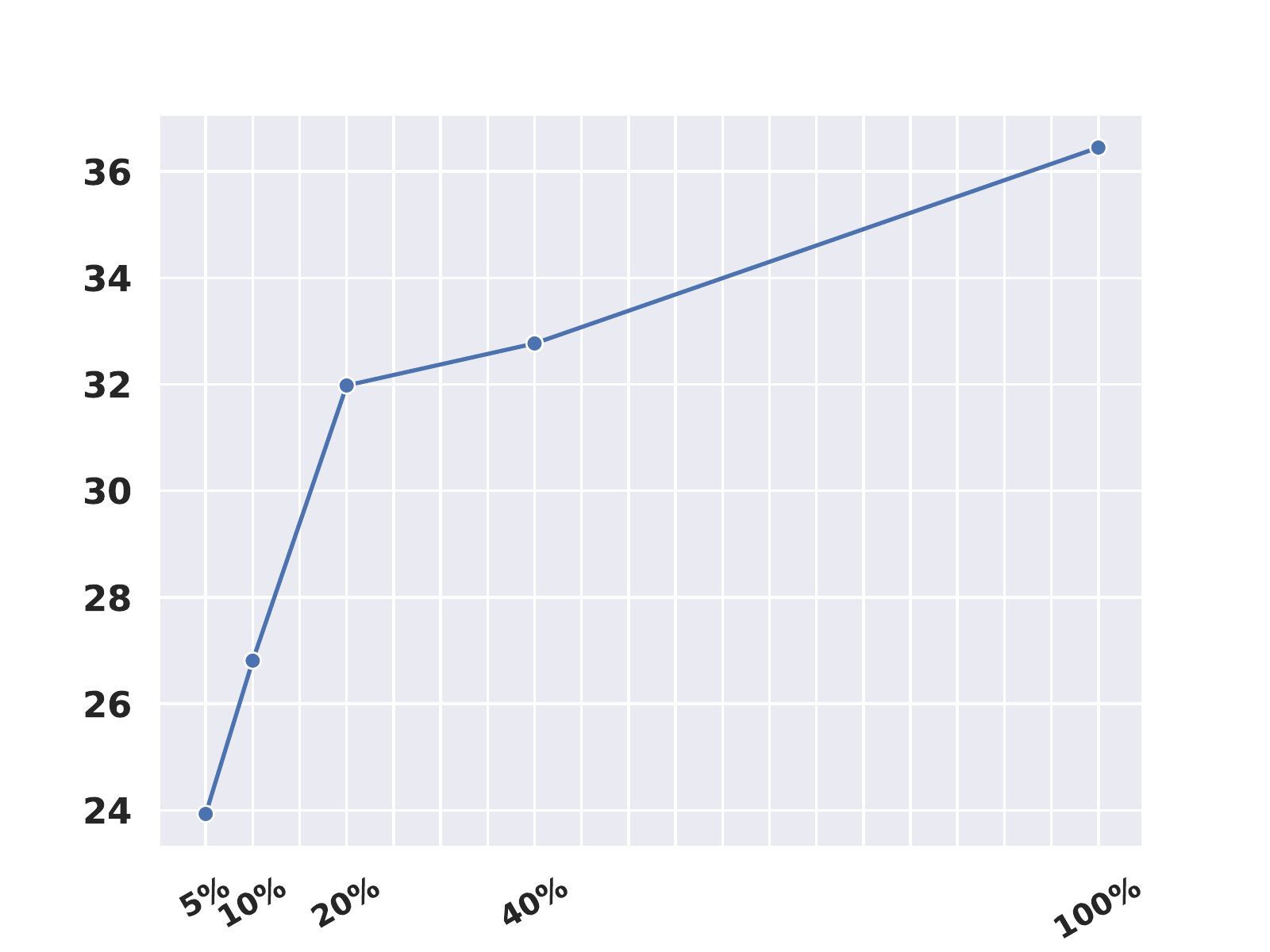}
\end{minipage}%
}
\caption{Results of TKK-Base as the amount of training data in the knowledge acquisition stage increases.}
\label{fig:KA}
\end{figure}

\begin{figure}[tb]
\centering
\subfigure[EM results on \dataset{Spider}]{
\centering
\begin{minipage}[t]{0.5\linewidth}
\centering
\includegraphics[width=1.4in]{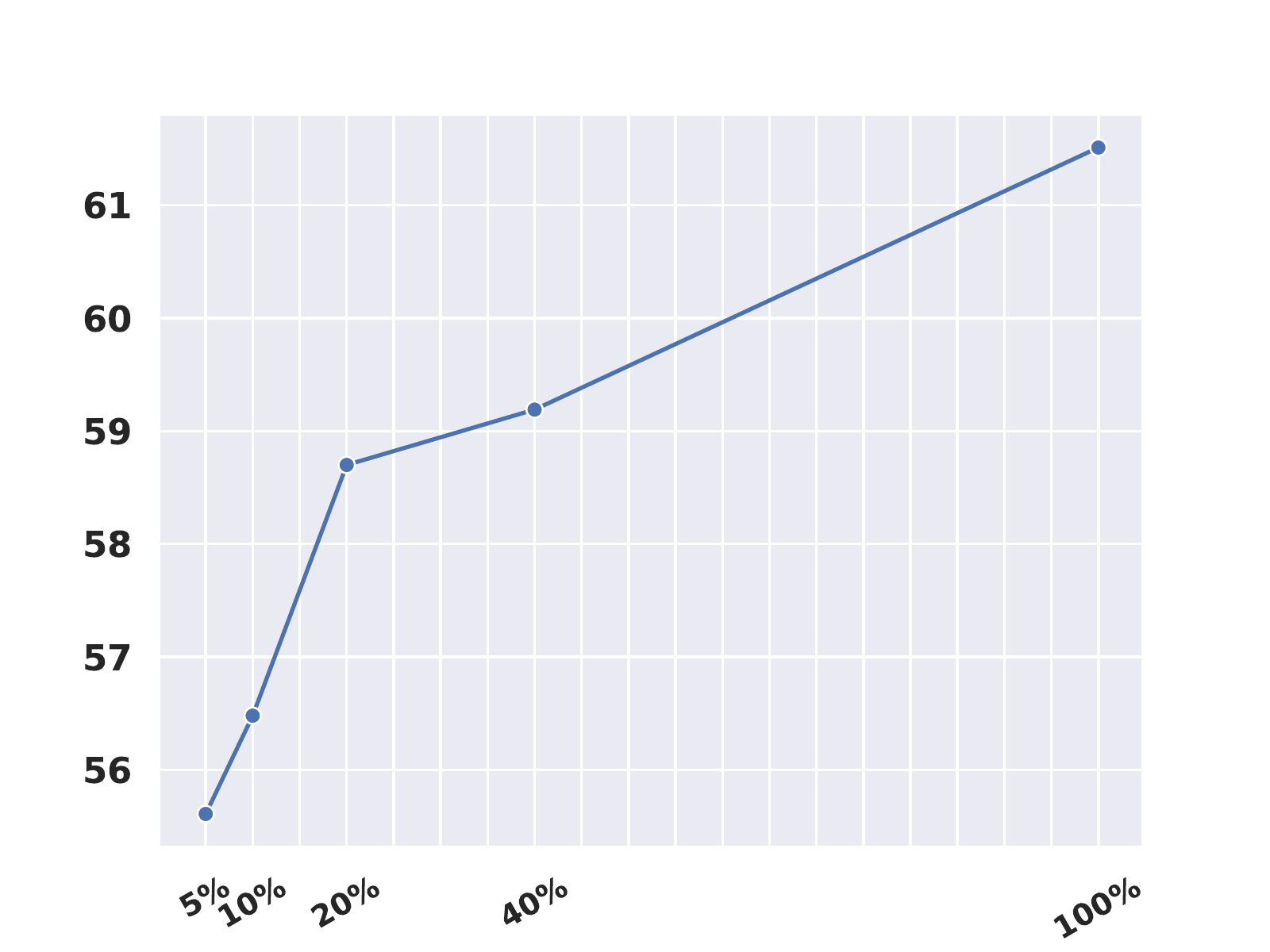}
\end{minipage}%
}%
\subfigure[QM results on \dataset{CoSQL}]{
\centering
\begin{minipage}[t]{0.5\linewidth}
\centering
\includegraphics[width=1.4in]{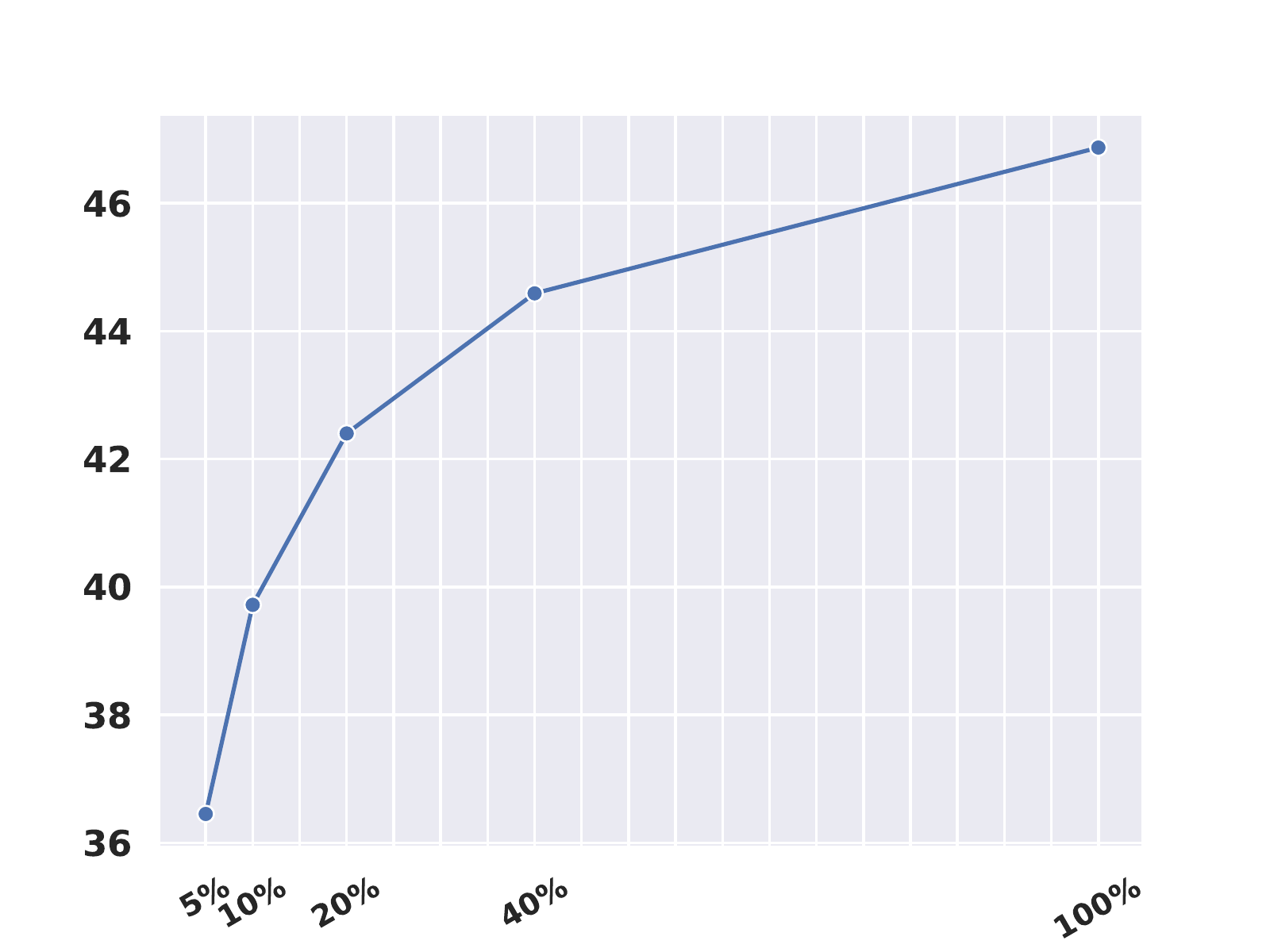}
\end{minipage}%
}
\caption{Results of TKK-Base as the amount of training data in the knowledge composition stage increases.}
\label{fig:KC}
\end{figure}

\textbf{Is each subtask necessary in the knowledge acquisition stage?} 
To quantify the contribution of each subtask, we examine the performance of the main task after removing a subtask for training in the knowledge acquisition stage. Table \ref{tab:ablation} shows the ablation results on \dataset{Spider} and \dataset{CoSQL}. Removing any subtask degrades the model's performance on the main task, indicating that all subtasks are necessary for the knowledge acquisition stage. We can see that the \sql{FROM} subtask has the largest impact on the performance. This is due to the mismatch between natural language expression and SQL syntax. User questions generally do not involve which tables to retrieve data from, while SQL requires specifying this. The \sql{FROM} subtask allows the model to learn the alignment between the question and the \sql{FROM} clause, thus alleviating the mismatch problem. Although other clauses are more or less mentioned in user questions, there are also alignment issues. Some previous work tackles this problem via designing intermediate representations \cite{irnet, natsql}. Our framework provides a new perspective. The effect of the \sql{SQL} subtask is less pronounced since the number of training examples of it is much smaller than that of the other subtasks. 
\begin{table}
\centering
\begin{tabular}{lcccc}
\hline
&  \multicolumn{2}{c}{\dataset{Spider}} & \multicolumn{2}{c}{\dataset{CoSQL}} \\
\textbf{Models} & \textbf{EM} & \textbf{EX} & \textbf{QM} & \textbf{IM} \\
\hline
TKK-Base & \textbf{61.5} & \textbf{64.2} & \textbf{46.9} & \textbf{17.8}\\
\quad w/o SELECT & 60.0 & 63.4 & 43.8 & 15.0 \\
\quad w/o FROM & 60.0 & 62.0 & 43.2 & 14.3 \\
\quad w/o WHERE & 60.0 & 63.4 & 44.6 & 16.7 \\
\quad w/o GHOL & 60.9 & 63.1 & 43.5 & 16.0  \\
\quad w/o SQL & 61.2 & 63.4 & 45.3 & 17.1 \\
\hline
\end{tabular}
\caption{The effect of subtasks.}
\label{tab:ablation}
\end{table}
\\
\textbf{How effective is knowledge acquisition?}
We want to investigate if adding training data in the knowledge acquisition stage will further improve the model's performance. To this end, we first take 5\%, 10\%, 20\%, 40\%, and 100\% of the data for constructing the training data in the knowledge acquisition stage and then use 5\% of the data for finetuning with the main task. The results on \dataset{Spider} and \dataset{CoSQL} are shown in Figure \ref{fig:KA}. As the amount of training data in the knowledge acquisition stage increases, the performance of TKK-Base improves significantly. This suggests that pre-training the model with large-scale subtask data will be beneficial to improving the model's performance.
\\
\\
\textbf{How effective is knowledge composition?}
To answer this, we use all data in the knowledge acquisition stage for training and then take 5\%, 10\%, 20\%, 40\%, and 100\% of the data for finetuning with the main task. As shown in Figure \ref{fig:KC}, training with more data in the knowledge composition stage is also helpful. Since only using subtasks for training loses the dependency information of different subtasks, knowledge composition helps the model to capture this information. Moreover, fine-tuning the main task with only 5\% data has already achieved 90\% and 78\% of the performance fine-tuned with all the data on \dataset{Spider} and \dataset{CoSQL}, respectively, showing that the model only needs a small amount of data to learn to tackle the main task.
\\
\\
\textbf{What is the model's performance in terms of different hardness levels?}
SQL queries in \dataset{Spider} can be divided into four hardness levels: easy, medium, hard, and extra hard \cite{spider}. Table \ref{tab:hardness} shows a  comparison between TKK and T5 regarding these four hardness levels. It can be seen that the performance improvement mainly comes from hard and extra hard examples. For example, TKK-Base and TKK-Large improve T5-Base and T5-Large by 16.3 and 10.9 absolute points on extra hard examples, respectively. 
By dividing the learning process into multiple stages, our framework dramatically improves the model's ability to handle complex queries, thus leading to better overall performance.
\begin{table}[tb]
\centering
\setlength\tabcolsep{3.5pt}
\begin{tabular}{lcccc}
\hline
\textbf{Models} & \textbf{Easy} & \textbf{Medium} & \textbf{Hard} & \textbf{Extra} \\
\hline
T5-Base  & 83.9 & 59.9 & 42.5 & 22.9 \\
T5-Large & 87.5 & 74.0 & 50.0 & 34.3 \\
\hline
TKK-Base & 83.9 & 63.0 & 47.1 & 39.2 \\
TKK-Large & 89.5 & 76.5 & 52.9 & 45.2 \\
\hline
\end{tabular}
\caption{EM results on \dataset{Spider} in terms of different hardness levels.}
\label{tab:hardness}
\end{table}
\\
\\
\textbf{Is TKK still effective in low-resource scenarios?} 
\begin{figure}[tb]
\centering
\subfigure[EM results on \dataset{Spider}]{
\centering
\begin{minipage}[t]{0.5\linewidth}
\centering
\includegraphics[width=1.4in]{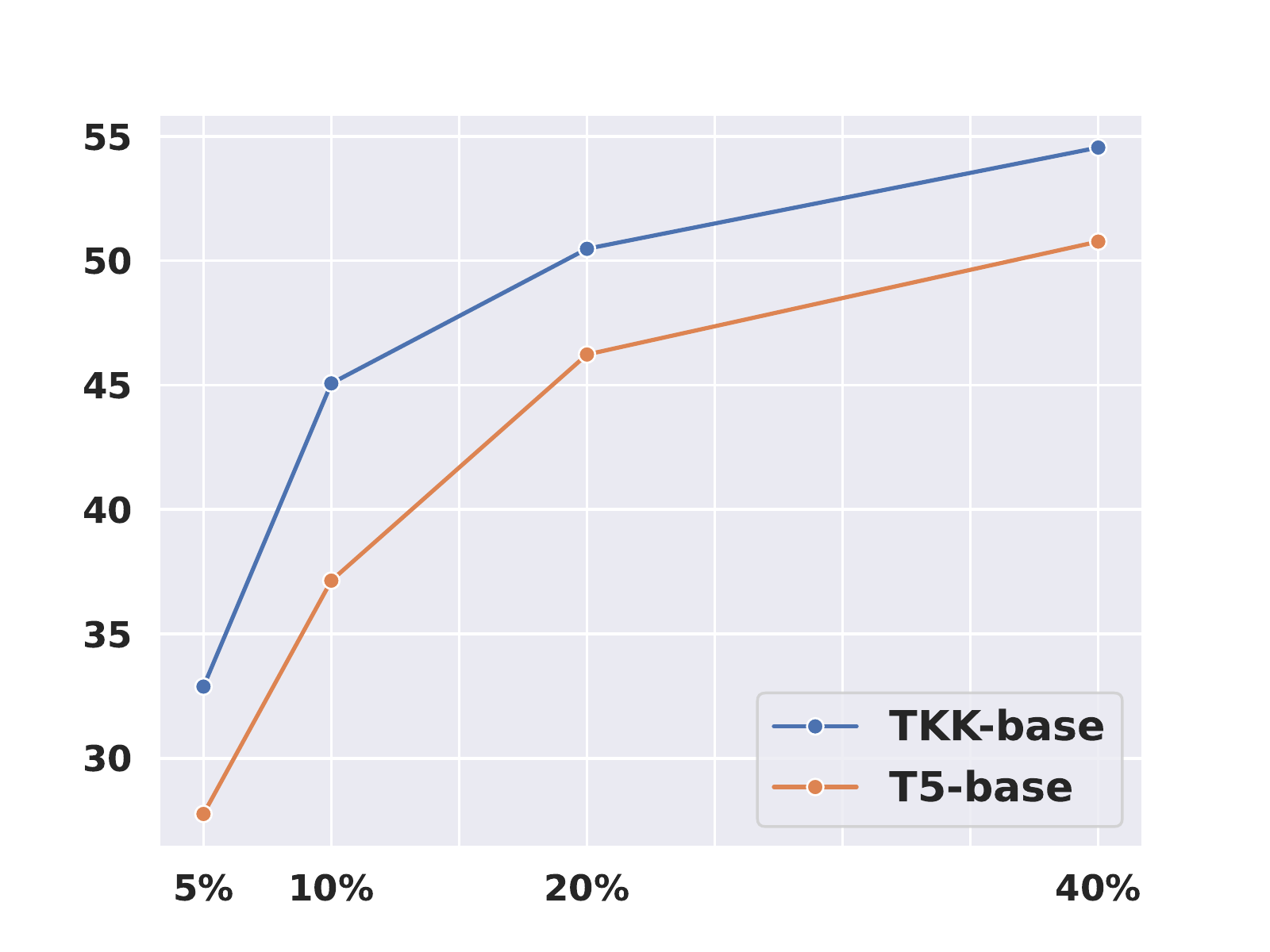}
\end{minipage}%
}%
\subfigure[QM results on \dataset{CoSQL}]{
\centering
\begin{minipage}[t]{0.5\linewidth}
\centering
\includegraphics[width=1.4in]{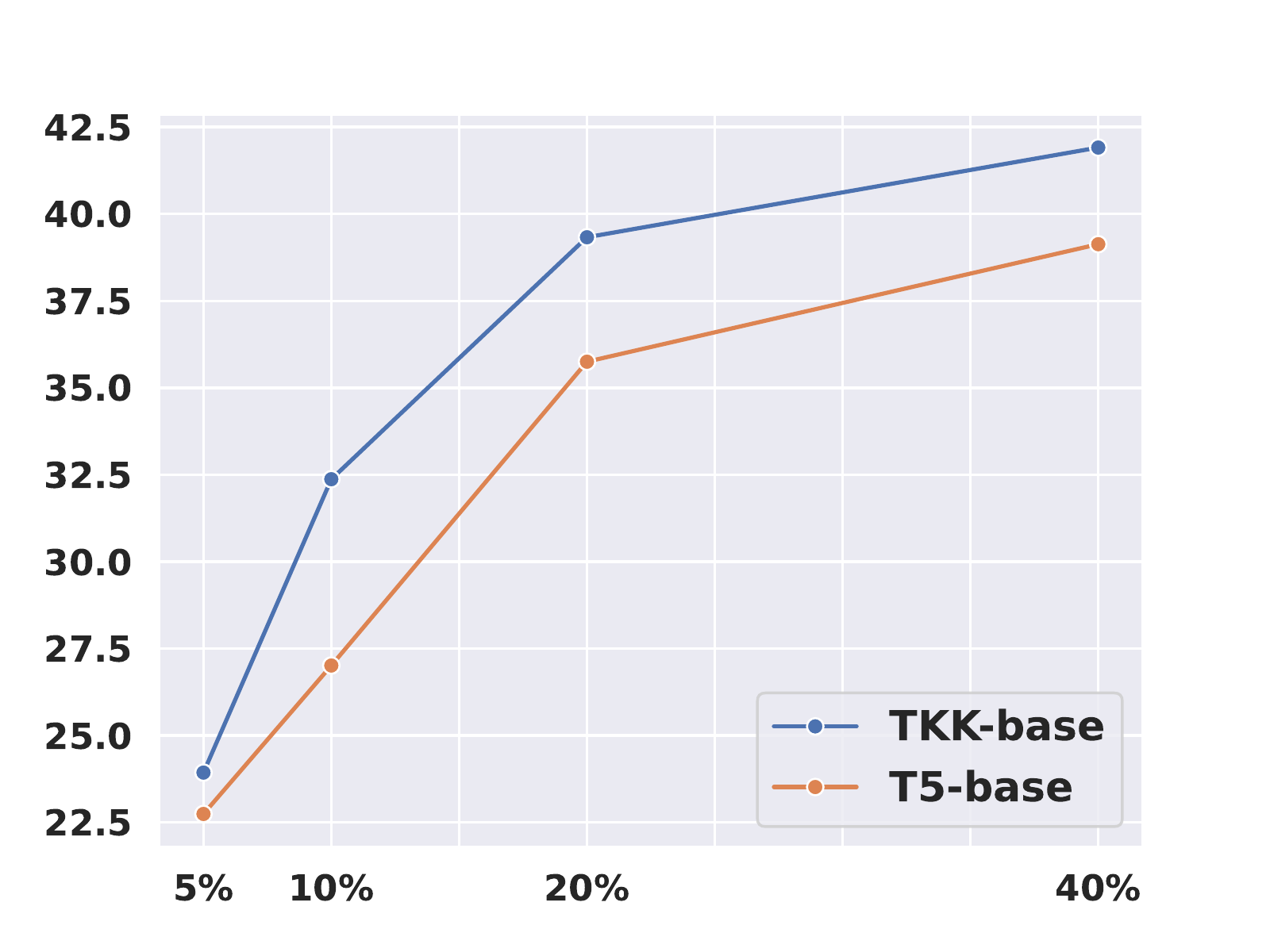}
\end{minipage}%
}
\caption{Results of T5-Base and TKK-Base in low-resource scenarios.}
\label{fig:low}
\end{figure}
Another perspective to study the model's generalization ability is to investigate its performance in the low-resource setting. To this end, we conduct experiments on \dataset{Spider} and \dataset{CoSQL}. For each dataset, we randomly shuffle the training set and then take 5\%, 10\%, 20\%, and 40\% of the data for training. The results of T5-Base and TKK-Base are shown in Figure \ref{fig:low}.  TKK-Base performs better than T5-Base no matter how much data is used for training, showing that our framework is also effective in low-resource scenarios.
\\
\\
\textbf{Case Study} We also conduct a case study to show that TKK makes fewer mistakes on SQL clauses and is more robust to synonym substitution compared with T5. Details are in Appendix \ref{case}.

\section{Related Work}

Most previous work aims to solve a particular generalization or robustness challenge for text-to-SQL parsing.  \citet{dong-lapata-2016-language} introduce a sequence-to-tree approach  for traditional i.i.d. datasets such as GeoQuery \cite{zelle1996learning}. \citet{syntaxsqlnet} propose a syntax tree network to tackle the zero-shot text-to-SQL problem. Later, various methods address the challenge from different perspectives such as improving schema linking \cite{ratsql, hui2021improving, qin2021sdcup, proton}, data augmentation \cite{yu2021grappa, wu-etal-2021-data}, or exploiting history information \cite{r2sql, hiesql}.  \citet{shaw-etal-2021-compositional} combines a grammar-based approach with T5 \cite{T5} to address the compositional generalization challenge.  \citet{deng-etal-2021-structure} develop a structure-grounded pre-training framework for improving the model's robustness against natural language variations. \citet{pi-etal-2022-towards} build an adversarial training example generation framework to bring the model better robustness against table perturbations.  However, the success of specialized architectures or training approaches on one challenge cannot easily transfer to others \cite{unlocking, furrer2020compositional}. Our TKK framework, for the first time, shows improvements in all the concerned challenging scenarios for text-to-SQL parsing. 

Our work is also related to the research of task decomposition in NLP \cite{unigdd, nye2022show, wies2022sub, wei2022chain, wang2022self}. For example, least-to-most prompting \cite{zhou2022least}, a method purely based on inference with a sufficiently large pre-trained language model, 
reduces a complex task into multiple subtasks and solves these subtasks sequentially. By comparison, TKK first learns to solve simpler subtasks and then the complex task. At inference time, the model directly tackles the complex task.

\section{Conclusion}

This paper proposes a general and effective TKK framework for text-to-SQL parsing, which has three stages: task decomposition, knowledge acquisition, and knowledge composition. TKK enhances the model's ability to acquire general SQL knowledge by dividing the learning process into multiple stages. Comprehensive evaluation on three levels of generalization, namely \textit{i.i.d.}, \textit{zero-shot}, and \textit{compositional}, and robustness demonstrates the effectiveness of our framework. 

\section*{Limitations}

Although our TKK framework is conceptually simple, it needs to decompose the task into multiple subtasks manually. It is not difficult to decompose the text-to-SQL parsing task due to the simplicity of SQL syntax. However, decomposing the complex graph structure such as Abstract Meaning Representation (AMR) is not straightforward. Therefore, a general strategy to automatically discover the meaningful substructure of the original task is needed. With such a strategy, our framework can be extended to broader research areas as long as the task can be decomposed into meaningful subtasks. We aim to address this limitation in our future work.


\bibliographystyle{acl_natbib}

\appendix

\section{Case Study}
\label{case}

Table \ref{tab:case} shows some real cases including questions and the SQL queries generated by T5-3B and TKK-3B. The first four cases are from the \dataset{Spider} dataset. It can be seen that TKK can produce correct SQL queries, while T5 makes mistakes on some clauses. The last two cases are from the \dataset{Spider-Syn} dataset. After synonym substitution, T5 is confused by synonyms and outputs some tables and columns that do not exist in the database, leading to incorrect SQL queries. By comparison, TKK can still identify correct tables and columns.

\begin{table*}
\centering
\begin{tabular}{l}
\hline
\multicolumn{1}{c}{\dataset{Spider}} \\
\hline
\textbf{Question:} What is the average GNP and total population in all nations whose government is \\  US territory? \\
\textbf{T5-3B:} SELECT avg(gnp), \red{avg}(population) FROM country WHERE governmentform = \\ "US Territory" \\
\textbf{TKK-3B:} SELECT avg(gnp), \blue{sum}(population) FROM country WHERE governmentform = \\ "US Territory" \\
\hline
\textbf{Question:} Which model of the car has the minimum horsepower? \\
\textbf{T5-3B:} SELECT model FROM cars\_data ORDER BY horsepower LIMIT 1 \\
\textbf{TKK-3B:} SELECT \blue{car\_names.}model FROM cars\_data \blue{JOIN car\_names on cars\_data.id =} \\ \blue{car\_names.makeid} ORDER BY horsepower LIMIT 1 \\
\hline
\textbf{Question:} What are the dog name, age and weight of the dogs that were abandoned? Note that \\ 1 stands  for yes,  and 0 stands for no in the tables. \\
\textbf{T5-3B:} SELECT name,  age,  weight FROM dogs WHERE abandoned\_yn  =  \red{0} \\
\textbf{TKK-3B:} SELECT name,  age,  weight FROM dogs WHERE abandoned\_yn  =  \blue{1} \\
\hline
\textbf{Question:} Return the different document ids along with the number of paragraphs corresponding \\ to each,  ordered by id. \\
\textbf{T5-3B:} SELECT document\_id,  count(*) FROM paragraphs GROUP BY document\_id \\ ORDER BY  \red{count(*)} \\
\textbf{TKK-3B:} SELECT document\_id,  count(*) FROM paragraphs GROUP BY document\_id \\ ORDER BY  \blue{document\_id} \\
\hline
\multicolumn{1}{c}{\dataset{Spider-Syn}} \\
\hline
\textbf{Question-O:} How many templates do we have? \\
\textbf{Question-S:} How many layout do we have? \\
\textbf{T5-3B:} SELECT count(*) FROM \red{layout} \\
\textbf{TKK-3B:} SELECT count(*) FROM \blue{templates} \\
\hline
\textbf{Question-O:} What is the year that had the most concerts? \\
\textbf{Question-S:} What is the time that had the most shows? \\
\textbf{T5-3B:} SELECT \red{time} FROM concert GROUP BY \red{time} ORDER BY count(*) desc LIMIT 1 \\
\textbf{TKK-3B:} SELECT \blue{year} FROM concert GROUP BY \blue{year} ORDER BY count(*) desc LIMIT 1 \\
\hline
\end{tabular}
\caption{Case study. Question-O is the question in the original \dataset{Spider} dataset. Question-S is the question in the \dataset{Spider-Syn} dataset, modified from Question-O using synonym substitution.}
\label{tab:case}
\end{table*}

\end{document}